\documentclass[conference]{IEEEtran}

\IEEEoverridecommandlockouts

\usepackage{cite}
\usepackage{amsmath,amssymb,amsfonts}
\usepackage{algorithmic}
\usepackage{algorithm}
\usepackage{comment}
\usepackage{graphicx}
\usepackage{caption}
\usepackage{subcaption}
\usepackage{graphics}
\usepackage{textcomp}
\usepackage{xcolor}
\usepackage{hyperref}
\usepackage{xcolor}
\hypersetup{
    colorlinks,
    linkcolor={blue!50!black},
    citecolor={blue!50!black},
    urlcolor={blue!80!black}
}

\def\BibTeX{{\rm B\kern-.05em{\sc i\kern-.025em b}\kern-.08em
    T\kern-.1667em\lower.7ex\hbox{E}\kern-.125emX}}
\begin{document}



\title{Fault Detection in Telecom Networks using \\ Bi-level Federated Graph Neural Networks
}

\author{\IEEEauthorblockN{Rémi Bourgerie}
\IEEEauthorblockA{KTH Royal Institute of Technology}
\IEEEauthorblockA{\texttt{remibo@kth.se}}
\and
\IEEEauthorblockN{Tahar Zanouda}
\IEEEauthorblockA{Global AI Accelerator, Ericsson AB.}
\IEEEauthorblockA{\texttt{tahar.zanouda@ericsson.com}}}

\maketitle

\thispagestyle{plain}
\pagestyle{plain}

\begin{abstract}

5G and Beyond Networks become increasingly complex and heterogeneous, with diversified and high requirements from a wide variety of emerging applications. 
The complexity and diversity of Telecom networks place an increasing strain on maintenance and operation efforts. Moreover, the strict security and privacy requirements present a challenge for mobile operators to leverage network data. To detect network faults, and mitigate future failures, prior work focused on leveraging traditional ML/DL methods to locate anomalies in networks. The current approaches, although powerful, do not consider the intertwined nature of embedded and software-intensive Radio Access Network systems.\\ 
In this paper, we propose a Bi-level Federated Graph Neural Network anomaly detection and diagnosis model that is able to detect anomalies in Telecom networks in a privacy-preserving manner,  while minimizing communication costs. Our method revolves around conceptualizing Telecom data as a bi-level temporal Graph Neural Networks. The first graph captures the interactions between different RAN nodes that are exposed to different deployment scenarios in the network, while each individual Radio Access Network node is further elaborated into its software (SW) execution graph. Additionally, we use Federated Learning to address privacy and security limitations. Furthermore, we study the performance of anomaly detection model under three settings: (1) Centralized (2) Federated Learning and (3) Personalized Federated Learning using real-world data from an operational network. Our comprehensive experiments showed that Personalized Federated Temporal Graph Neural Networks method outperforms the most commonly used techniques for Anomaly Detection.

\end{abstract}

\begin{IEEEkeywords}
TelcoAI, Federated Learning, Graph Neural Networks, Graph-based Federated Learning, Anomaly Detection.

\end{IEEEkeywords}

\section{Introduction}
In today’s connected world, Telecom providers strive to ensure reliable and uninterrupted services. Undoubtedly, the massive increase in data demand \cite{EricMobilityReport} introduces new challenges in how Telecom providers plan and maintain their networks. Large Telecom networks are typically heterogeneous, and continuously evolving. Telecom networks comprise \textit{Radio Access Networks (RAN)} products supported by complex software-intensive embedded systems. Software (SW) Modules in RAN are designed to handle logic and compute functionalities. Due to the nature of Telecom networks, several functionalities (e.g., Handover) depend on neighbouring RAN nodes in the network. The diversity and the intertwined complexity of technologies powering RAN present a challenge for mobile operators, the network can experience SW failures, or SW faults, due to various reasons. The failure in the network can lead to the unavailability of network services, and the occurrences of such failures could cause significant revenue loss. Debugging Telecom system faults can be a tedious process, the challenge compounds when considering that a RAN node is not functioning in isolation in the network. Another challenge is about data handling, meaning the collection and transport of data \cite{EricDataIngestion} needed for training models. The data collection process requires computational, storage, and transport resources. As RAN resources are constrained, these resources can otherwise be used for RAN functionalities (e.g. serving traffic, etc.).  

In order to mitigate faults in networks, Telecom SW vendors collect SW performance data to enable instant feedback and alerts for any aberrant behaviour in the system. This data comes in the form of SW checkpoints, implemented by SW vendors, to monitor the execution of SW to further locate faults easily. SW checkpoints can be checked using event counters over time. Many efforts have been used to leverage AI/ML to simplify troubleshooting efforts. Traditional fault detection methods typically learn patterns from historical RAN data over time to detect anomalies. However, the main challenge we faced is the limitation of using the intertwined (1) hierarchical and modular nature of SW execution, and (2) the connection between nodes in telecom networks that are geographically deployed to provide coverage.

Due to the sensitive nature of Telecom data, it is a challenge to use, store and process data. In addition, Telecom network data is collected directly from RAN nodes \cite{EricDataIngestion}, and typically store data in a central server to train machine learning models. With this regard, Federated Learning (FL) \cite{Mammen2021} provides a promising framework as it may allow the Telecom industry to train models using isolated data from RAN nodes, and even from multi-operators, without the need to share the raw data between data providers. Current 3GPP standardization efforts expect the use of Federated Learning in Telecom AI in the near future \cite{lin2023artificial}. As the same SW is used to power RAN nodes, leveraging the topology of how RAN nodes are deployed to combine the knowledge of RAN nodes exposed to similar scenarios, can potentially help reduce faults in the network.

In this paper, we propose a bi-level Federated Graph Neural Network approach that aims to identify anomalies within the Telecom network while minimizing communication costs. To the best of our knowledge, this work is the first to consider a bi-level Federated Graph Neural Network to combine SW execution flow with Telecom NW dependencies to detect faults in the network. The following are the contributions:
\begin{itemize}
    \item We leverage Telecom knowledge expertise to conceptualize and combine multi-dimensional RAN data with temporal SW execution information as a bi-level temporal graph. 
    \item We evaluate several models and strategies to detect anomalies in 4G/5G software using the following three setups:

    \begin{enumerate}
        \item \textit{Centralized Temporal Graph Neural Networks Model}: we present a centralized model designed to identify anomalies in the 4G/5G Telecom data.
        \item \textit{Federated Temporal Graph Neural Networks Model}: we introduce Federated Learning (FL) as a privacy-preserving training mechanism for fault detection models.
        \item \textit{Personalized Federated Temporal Graph Neural Networks Model}: We introduce an innovative aggregation technique referred to as FedGraph. 
    \end{enumerate}
    
    \item To demonstrate the proposed method's performance in anomaly detection tasks, we manually simulated anomalous scenarios with the help of domain experts and treated it as a true anomaly.

\end{itemize}

The paper is organized as follows. Section II reviews the state-of-the-art. Data is discussed in section III. Section IV mathematically formulates the problem, and outlines the proposed method. Section V provides details on evaluation metrics. Experimental results and analysis are introduced in Section VI and discussion in Section VII. Finally, Section VIII concludes the paper.

\section{Related work}

\subsection{Anomaly detection on multivariate time series}

Anomaly detection \cite{chalapathy2019deep} refers to the identification of rare items, events or observations which deviate significantly from the dataset's normal behaviour.

Methods for anomaly detection based on clustering techniques, or Decision Trees  \cite{chalapathy2019deep} have proven effective in many cases thanks to their rapid convergence and easy implementation. However, they may not perform well on multivariate time series since they consider time series as simple data vectors and as such fail to capture the time-dependent properties.
The most popular anomaly detection methods are model-based \cite{blazquez2020}. methods based on Recurrent Neural Networks \cite{rumelhart1986learning}, like GRU \cite{cho2014properties} and LSTM \cite{hochreiter1997long} became widely popular for modeling time series. A model is trained to model the normal behaviour of the input data through a reconstruction task. Then the reconstruction is compared with the real data and an anomaly score is computed for each data point based on Outlier Detection methods. These methods \cite{blazquez2020} identify points that differ significantly from their expected value. The most simple approaches, like the z-score \cite{altman1968financial}, apply a threshold on the reconstruction error to identify anomalies. Alternatively, methods based on t-test \cite{beaver1966financial}, test some statistical hypothesis on the reconstruction error to identify outliers. In the case of the Extreme Student Deviation (ESD) test \cite{mehrang2015outlier}, the outlier hypothesis is recursively tested on the most extreme values until no more outliers are identified. Some \cite{hochenbaum2017automatic} also propose to use estimators robust to outliers in the ESD test.

\subsection{Temporal Graph Neural Networks}

Recently, GNNs have been widely used for different tasks across different fields. A number of methods have been proposed that combine or extend these techniques for Temporal Graph data. Seo \& al. \cite{Seo2016} developed several Recurrent Graph Neural Networks. They are built out of a Recurrent Unit stacked on top of a Graph Convolutional Network, which means that a Recurrent Network is applied to the node embeddings obtained from a GNN at each time step. Popular architectures are the natural derivation of GRUs (GraphConvGRU) or LSTMs (GraphConvLSTM) \cite{seo2018structured}. Moreover, Attention-based Temporal GNNs have been proposed like A3T-GCN \cite{Zhu2020}.
Due to the spatial and distributed nature of RAN data, employing GNN capture topological dynamics of the system \cite{shen2022graph}. Early work has investigated the blossom of GNNs for fault Detection \cite{yen2022graph, yu2023flam, qiancellular}. The approaches, although powerful, do not consider the intertwined nature of complex software-intensive RAN systems with inter-metric dependency of internal SW execution flow, the approaches leverage the inter-metric dependency of Key Performance Indicators (KPI).

\subsection{Federated Learning}

Federated Learning, first referred to as Federated Optimization, is a concept introduced by Google in 2016 \cite{Konecny2016}. The most popular FL strategy referred to as Federated Average was proposed by MacMahan \& al \cite{McMahan2016}. They use the Stochastic Gradient Descent algorithm to solve the respective local training tasks and then aggregate the models by averaging the local model weights. While training on highly heterogeneous data (non-iid), the model for the average client is prone to drift towards common behaviour among clients. Tan \& al. \cite{Tan2021} motivate the use of personalized FL models to address the drawbacks of building one common model for every client. Furthermore, graph-based aggregation opened the way for leveraging graph topology between clients. Chen \& al. \cite{Chen2022} proposed to build a fully connected graph and then set the edge values based on similarity scores between the local models. The local models are then aggregated while performing message passing on the graph and taking the local models' weights as node features.
Xing \& al. \cite{xing2022big} propose BiG-Fed, a bilevel optimisation method for graph-aided Federated Learning. Two training loops are used asynchronously, the upper training optimizes the aggregation task, whereas the inner loop optimizes the training task on local data.

Due to the sensitive nature of Telecom data, Federated Learning has been adopted as a viable solution \cite{letaief2019roadmap} to handle data for current and next-generation networks.

\section{Data}
\label{sec:dataCollection}

A telecom operational network consists of interconnected RAN nodes that provide network coverage. Each RAN node (e.g., gNB, eNB, etc) is running a SW that can be monitored to assess the behavior of the system. Data is being collected continuously \cite{EricDataIngestion}. w

\subsection{Data Sampling}

As Telecom operators strive to provide reliable service, faults are very rare in the network. Telecom SW undergoes extensive tests before being deployed. Given the diversity of deployment scenarios, ensuring that such a method is assessed and further developed may require an enormous flow of training and testing data. For experimental purposes, this work uses data collected from a representative subset of RAN nodes to evaluate a diverse subset of development environments as depicted in table \ref{tab:ts_collected}. However, the data collected does not come in an annotated form.

\begin{table}[h!]
    \centering
    \begin{tabular}{|l|l|}
    \hline
    Number of cells             & 67  among which \\ & airport area: 12 \\ & downtown area: 29 \\ & rural area: 26\\ \hline
    Collection date             &  2022-10-01 00:00:00 -\\ & 2022:11:10 23:59:59\\  \hline
    Collection frequency        & every 15 mins                               \\ \hline
    Length of time series       & 3072                                     \\ \hline
    Number of time-series measurements          & 24                                         \\ \hline
    Total amount of data points & 3 millions                                        \\ \hline
    \end{tabular}
    \caption{Description of the time Series data collected}
    \label{tab:ts_collected}
\end{table}

\subsection{Evaluation Dataset}
\label{sec:data_annotation}
In order to evaluate our work, we have manually added anomalous data with the help of domain experts into the test data and treated it as a true anomaly. The algorithm \ref{alg:data_annotation} has been developed with a SW domain expert to craft failure scenarios. Reproducing a realistic scenario of SW faults involves designing an annotation algorithm focusing on:
\begin{itemize}
    \item Introducing anomalies on checkwpoints where anomalies are more likely to occur.
    \item A realistic number of anomalies on abnormal cells.
    \item A realistic propagation of the anomaly to other counters.
    \item A realistic magnitude of the anomalies.
    \item A realistic location of the abnormal cells.
\end{itemize}

\begin{algorithm}
\small
\begin{algorithmic}
\REQUIRE{
- Given $N$ RAN nodes in the network $\{C_{1},\ldots, C_{N}\}$   \\ 
- Given $\{ \{X_{1,1}^{t}, \ldots, X_{1,K}^{t}\}, \ldots, \{X_{N,1}^{t}, \ldots, X_{N,K}^{t}\}\} $ of $K$ time-series measurements for each RAN node\\
- Given $p$ an anomaly probability $p\in [0,1] $\\
- Given a RAN node sampling rule \\
- $A: \{s_1, \ldots , s_K\} \longrightarrow \mathbb{R}  $ representing the artificial degradation amplitude for each time-series measurement. 
}
    \FOR{each RAN node $C_i$}
        \STATE mark $C_i$ as abnormal or normal based on the RAN node sampling rule
        \IF{$C_i$ is abnormal}
            \FOR{each time-series $X_{i,k}^{t} \in \{X_{i,1}^{t}, \ldots, X_{i,K}^{t}\}$}
                \STATE sample random points $D \in \mathcal{P}(X_{i,k}^{t})$ in $X_{i,k}^{t}$ based on a uniform distribution of mean p
                \FOR{each point d in D}
                    \STATE - introduce and label d as an artificial anomaly based on anomalous case scenario and amplitude $A(s_k)$
                    \IF{propagation to dependent time-series measurements is enabled}
                        \STATE - propagate the same degradation to the relevant time-series measurements 
                        \STATE - mark the degraded points in the propagated time-series measurements as anomalies
                    \ENDIF
                \ENDFOR
            \ENDFOR
        \ENDIF
    \ENDFOR
\end{algorithmic}
\caption{Data annotation}\label{alg:data_annotation}
\end{algorithm}

\begin{table}[]
\begin{tabular}{|l|l|l|l|}
\hline
dataset & degraded cells           & \begin{tabular}[c]{@{}l@{}} \% of anomalies\\ introduced\end{tabular} &  \begin{tabular}[c]{@{}l@{}}with propagation\\ of anomalies\end{tabular} \\ \hline
0       & 0                        & 0               & -  \\ \hline
1       & airport area (12) & 0.01            & No           \\ \hline
2       & airport area (12) & 0.01      & Yes             \\ \hline
\end{tabular}
\caption{Characteristics of the dataset annotated. }
{\addvspace{-1\baselineskip}}
\label{tab:dataset}
\end{table}

The characteristics of the datasets annotated and later used for evaluation are described in table \ref{tab:dataset}. Dataset 0 represents the data collected without any annotation.

The data was first normalized using a Robust Scaler, to avoid any bias in the pre-processing introduced by any potential anomalies. The data has then been split with a sliding time window into batches containing the past sequence of the time series and the next sequence to predict, in a fashion that respects training temporal prediction models. For efficient inference during model training and evaluation, we employed PyTorch Geometric's mini-batching technique as detailed in \cite{fey2019fast}. Table \ref{tab:batched_data} displays the results of this data splitting and batching process.
\begin{table}[!htb]
\centering
\begin{tabular}{|l|l|l|}
\hline
Attribute   & Dimension                                                  & Description                                 \\ \hline
$x$           & [n\_nodes x batch\_size,  & nodes features of the \\ & node\_features\_dim, history] & past sequence \\ \hline
$y$           & [n\_nodes x batch\_size, & nodes features of the   \\ &  node\_features\_dim, horizon] & next sequence  \\ \hline
$e_i$ & {[}2, n\_edges{]}                                          & edges index                           \\ \hline
$e_a$  & {[}n\_edges{]}                                             & edges attributes                       \\ \hline
b       & {[}n\_nodes x batch\_size{]}                               & index of samples\\ & & in the batch        \\ \hline
\end{tabular}
\caption{Description of a batch of data used for the reconstruction task}
{\addvspace{-1\baselineskip}}
\label{tab:batched_data}
\end{table}

\section{Methods}

\subsection{Problem Formulation}

\noindent\textbf{Multivariate Time-Series:}
We consider a multivariate time-series of length $T$:
\[(\mathbf{p}^t)_{t=1}^{T} = (p_1^t, \ldots, p_K^t)_{t=1}^{T}\]
where each data point $p_k^t\in \mathbb{R}$ is collected at timestamp $t$ for the SW procedure $p_k$.

\noindent\textbf{Software (SW) Temporal Graph:}
Using the SW execution graph as edge knowledge, we structure these multivariate time series as a temporal graph, consisting of a set of $K$ vertices $\mathcal{V}$ and edges $\mathcal{E}$:
\[(\mathcal{G}_{\text{SW}}^t)_{t=1}^{T} = (\{\mathcal{V}^t, \mathcal{E}\})_{t=1}^{T} = (\{\mathbf{x}^t, \mathcal{E}\})_{t=1}^{T}\]
where $\mathcal{V}^t=\{x_1^t, \ldots, x_k^t\}$ represents the vertex features corresponding to the multivariate time series $(\mathbf{x}^t)_{t}$ at time step $t$, and $(\mathcal{E})_t=\mathcal{E}$ is the static edge set of the SW execution graph at time step $t$.

\noindent\textbf{Anomaly Detection:}
Given a temporal graph $(\mathcal{G}^t)_{t}$=$(\mathcal{V}^t, \mathcal{E}^t)_t$ as input with a time length of $T$, we aim to predict:
\[(\mathbf{y}^t)_{t=1}^{T} = \{y_1^t, \ldots, y_K^t\}_{t=1}^{T}\]
where $y_k^t \in \{0, 1\}$ denotes whether the data point at the timestamp of the node $x_k^t$ is anomalous ($1$ denotes an anomalous data point).

\noindent\textbf{Network (NW) Graph:}
The network-level graph consists of $N$ telecom RAN nodes evolving through $I$ iteration rounds:
\[(\mathcal{G}_{\text{NW}}^i)_{i=1}^{I}=(\{\mathcal{S}^i, \mathcal{R}^i\})_{i=1}^{I}=(\{\mathcal{S}, \mathcal{R}^{i}\})_{i=1}^{I}\]
where $S=\{S_1, \ldots, S_N\}$ represents the telecom RAN nodes, and $\mathcal{R}^i$ represents the relational links at iteration $i$ between the RAN nodes.

\subsection{Method Overview}
\label{sec:MethodOverview}

In this section, we describe a fault detection system using Bi-Level Federated Graph Neural Networks. Our method comprises two stages: 

\textbf{\textit{Stage I:}} Conceptualize and combine multi-dimensional RAN data with temporal SW execution information as a bi-level temporal graph. This step requires building (1) SW execution flow, and (2) Telecom Network Topology.

\textbf{\textit{Stage II:}} Design and implement Fault Detection Model. In this stage, we evaluate several models and strategies to detect anomalies in 4G/5G networks. In this stage, we evaluate the performance of the model under (1) Centralized, (2) Federated Learning and (3) Personalized Federated Learning settings using real-world data from an operational network.

\begin{figure}[H]
    \centering
    \includegraphics[width = 0.99\linewidth]{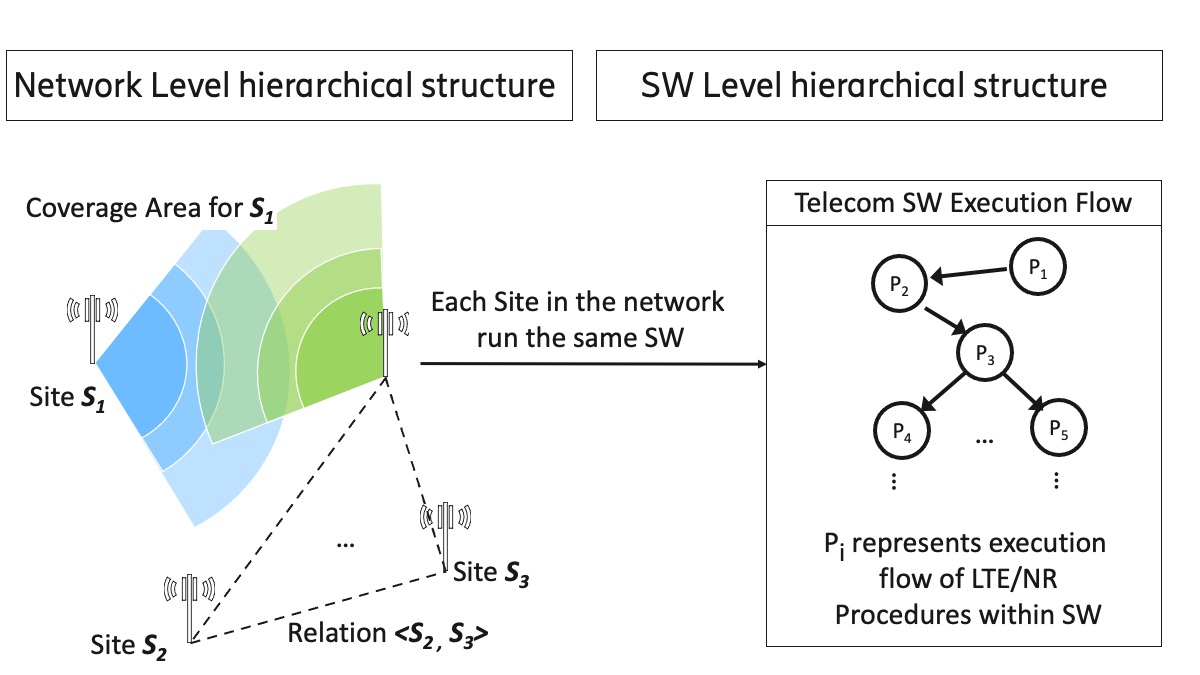}
    \caption{Illustration of bi-level graph used in the method}
    
    \label{fig:teaser}
\end{figure}

\textbf{\textit{Stage I: Encoding Telecom Network \& SW Execution data using Bi-level Temporal Graph}} 

A Telecom operational network consists of interconnected RAN nodes located in different regions. Each of these RAN nodes consists of cells that provide 4G/5G network coverage in a coverage geographic area. The topology graph can be constructed using mobility relations, transport relations (e.g., X2) or using spatial proximity.

\begin{figure}[htb]
\centering
  \includegraphics[width=\linewidth]{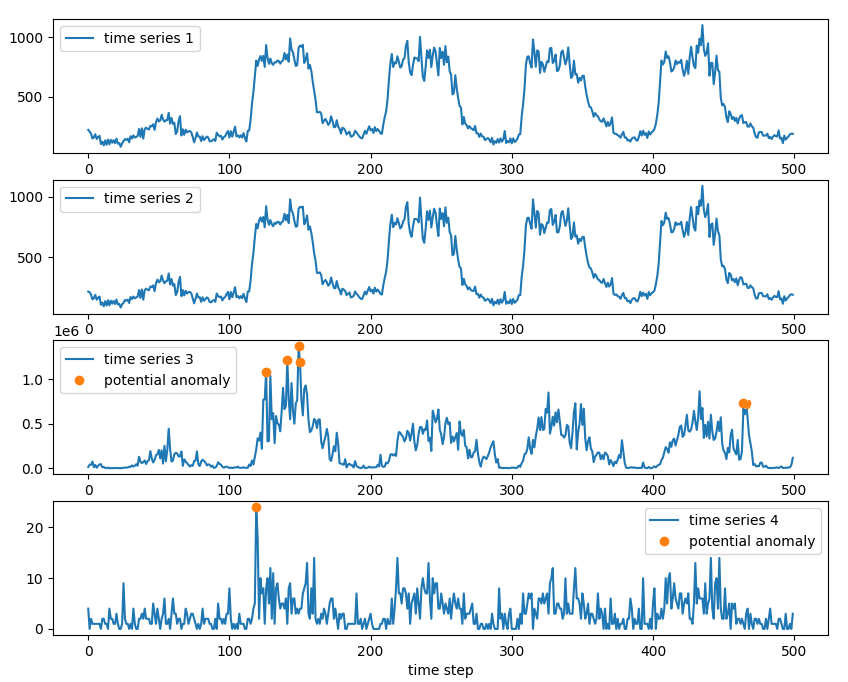}
  \caption{SW performance data}
  {\addvspace{-1\baselineskip}}
  \label{fig:performance_counters}
\end{figure}

Each RAN node in the network can execute the same SW. SW Execution Flow can be monitored using SW checkpoints. Each SW checkpoint can be monitored over time as illustrated in Figure \ref{fig:performance_counters}. Since RAN SW modules are complex, this is an attempt to build SW execution flow. This work characterized SW dependencies using procedures executed by LTE/NR user equipment (UE) and the various network elements to provide the services requested by the UE. Such procedures are often standardized as part of 3GPP \cite{lin2023artificial} standardisation.

There have been many proposals on how to semantically represent wireless data \cite{chaccour2022less}. In this work, we represent Telecom Network \& SW Execution data using a bi-level graph, as illustrated in fig \ref{fig:teaser}:
\begin{itemize}
    \item The first graph $G_{NW}$ represents a topology between the different RAN nodes.
    \item The second graph $G_{SW}$ represents the execution graph of a SW procedure being executed inside each RAN node. The links of this graph and the nodes are static and common to every RAN site. Only the node attributes evolve through time
\end{itemize}

Using this data structure, we manage to capture the dependencies between multivariate time series SW event data, with the intertwined hierarchical nature of Telecom networks. 
Our approach considers the SW execution graph $G_{SW}$ to structure the event data using relational links. Leveraging this graph-structured data is very powerful since promising machine learning techniques like Graph Neural Networks (GNNs) \cite{wu2020comprehensive} opened the way for the exploitation of graph-connected data.\\

\textbf{\textit{Stage II: Developing Fault Detection Mothod}} 

\label{sec:AnomalyDetection}

We evaluate several models and strategies to detect anomalies in 4G/5G software using the following three setups:

\begin{enumerate}
    \item \textit{Centralized Temporal Graph Neural Networks Model}
    \item \textit{Federated Temporal Graph Neural Networks Model}
    \item \textit{Personalized Federated Temporal Graph Neural Networks Model}
\end{enumerate}

In what follows, we describe each model architecture and evaluation process. 
\\
\subsubsection{\textbf{Centralized Temporal Graph Neural Networks Model}}

\textbf{\textit{Phase 1 - Input Reconstruction}} 

We propose an anomaly detection model using the reconstruction model which aims at predicting for every batch of data the next sequence $\hat{y}$, based on the input features $x$ and the edge features ($e\_i$ and $e\_a$):
\[
\text{arg}\min_{\omega} \sum_{(x, y, e_i, e_a, b) \in \mathcal{B}} L(y-f(\omega, x, e_i, e_a, b))
\]
with $L$: a distance function, $f$: the reconstruction model, $\omega$: the model parameters.

The reconstruction model comprises a Recurrent Graph Neural Network responsible for generating embeddings. This GNN is augmented with a Rectified Linear Unit (ReLU) layer to introduce non-linearity and further enhanced with a linear layer to perform next-step forecasting.

\noindent\textbf{\textit{Phase 2 - Anomaly Detection Function}} 

Two outlier detection methods, the Z-score and the ESD test, are employed together due to their superior performance on different signal types. The Z-score, a threshold-based approach, is preferred for counters with expected obvious anomalies, such as abrupt peaks. Conversely, the ESD test is applied to counters susceptible to anomalies of varying magnitudes. Consequently, the choice of outlier detection method is determined individually for each signal, with only the most effective method retained.

\subsubsection{\textbf{Federated Temporal Graph Neural Networks Model}}. \\
\label{sec:GraphFL}
We outline the Federated Learning strategy, as illustrated in Fig. \ref{fig:FL_process}, encompassing:
\begin{figure}
\centering
  \includegraphics[width=\linewidth]{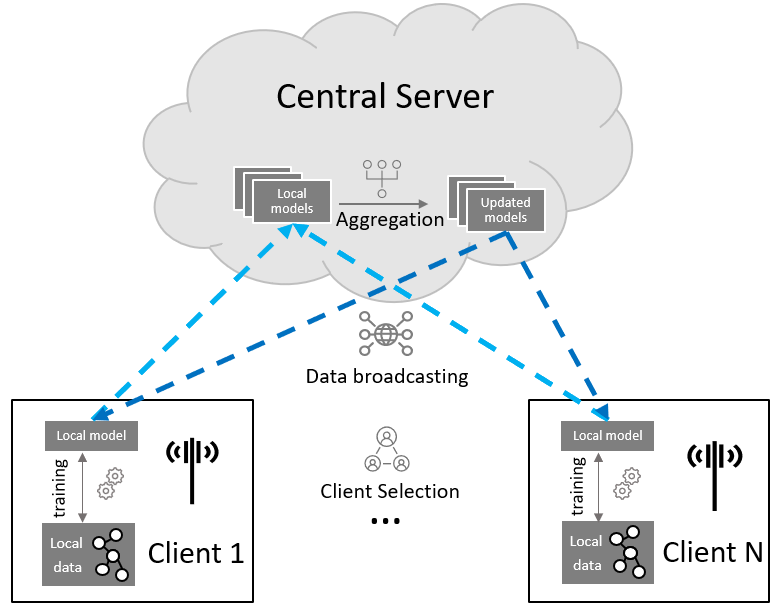}
  \caption{Illustration of a general Federated Learning process}
  
  {\addvspace{-1\baselineskip}}
  \label{fig:FL_process}
\end{figure}
\begin{itemize}
    \item Client selection criteria.
    \item Data exchange protocol between the server and clients.
    \item Local training procedures.
    \item Aggregation mechanism.
\end{itemize}

 All Federated Learning implementations discussed herein were implemented using Pytorch \footnote{https://pytorch.org/} and Flower \footnote{https://flower.dev/}.

All clients are sampled for local training. The data communicated between the server and the clients include:
\begin{itemize}
    \item From the clients to the server: This includes the local model weights and performance metrics, such as loss, precision, recall, and F1 score, calculated on their respective local data.
    \item From the server to the client: This includes the personalized model weights specific to each client and the global model weights.
\end{itemize}

In local training, two approaches are employed: MSE loss and MSE regularized loss, which is defined as follows: $ MSE_{reg} = MSE + \lambda(\omega - \omega^*)^2$
where $\lambda$ represents a regularization factor, $\omega$ the local model's weights and $\omega^*$ the global model's weights.

The FL vanilla aggregation rule FedAvg (defined in equation \ref{eq:FedAvg}), consists in averaging all the local model weights (of size $N_{LM}$) by individual clients to create a unified global model. This global model is distributed back to all clients for the subsequent round.
\begin{equation}
    FedAvg :\begin{cases}
        \omega^*_{global} = \sum_{S_i \in \{S_1, \ldots, S_N\}} {\omega}[i] / N \\
        {{\omega}[i]} = \omega^*_{global}  \text{ for } S_i \in \{S_1, \ldots, S_N\}
        
    \end{cases}
    \label{eq:FedAvg}
\end{equation}

\noindent where $\omega\in \mathbb{R}^{N\text{x}N_{LM}}$ denotes the local model weight matrix received by the server, $\omega^*\in \mathbb{R}^{N\text{x}N_{LM}}$ the local model weight matrix send to the client for the upcoming FL round and $\omega^*_{global}\mathbb{R}^{N_{LM}}$: the weights of the global model stored on the central server.

\subsubsection{\textbf{Personalized Federated Temporal Graph Neural Networks Model}}. \\
\label{sec:GraphFL}
\begin{equation}
    FedGraph :\begin{cases}
        \textit{Graph Definition:}\\
        \text{Let graph ${\mathcal{G}_{NW}}^{i}=\{\mathcal{S}, \mathcal{R}^{i}\}$}\\ 
        \text{with} \begin{cases}
        \text{$\mathcal{S}$} = \{S_1, ..., S_{N}\},\\
        \text{$\mathcal{R}^{i}$} [j, k] = 1 \text{ for $(S_j, S_k)$}\in\text{$\mathcal{S}$}^2
        \end{cases}\\
        \textit{Relation matrix tuning:}\\
        \mathcal{R}^i[j, k] = sim(\omega[j], \omega[k]) \text{ for $(S_j, S_k)$}\in\text{$\mathcal{S}$}^2 \\
        \textit{Message Passing: } 
        \boldsymbol{\omega^{*}} = {\mathcal{G}_{NW}}^{i}[\boldsymbol{\omega}] \\
        \textit{Global model: }
        \omega^{*}_{global} = \sum_{S_i}\boldsymbol{\omega^{*}}[i] / N
    \end{cases}
    \label{eq:FedGraph}
\end{equation}
where with the same notationsa as in Equation \ref{eq:FedAvg}$ \mathcal{S}$ denotes the vertices set, $\mathcal{R}^{i}\in \mathbb{R}^{N\text{x}N}$ the adjacency matrix at FL round $i$, $sim: \mathbb{R}^{N_{LM}} \text{x} \mathbb{R}^{N_{LM}} \longrightarrow [0, 1] $ a similarity function. In this work we use the cosine function as a similarity function: 
\[ cos(X, Y) = \frac{X.Y^T}{||X||_{2}. ||Y||_{2}}\]

In an effort to provide more personalized models, we introduce \textit{FedGraph} (defined in equation \ref{eq:FedGraph}) as an aggregation method. FedGraph establishes for every FL round $i$ a complete graph $\mathcal{G}_{SW}^{i}$ connecting all clients, associating each RAN node with its corresponding local model weights as node attributes. We further assign edge weights by computing a similarity score between the models of connected RAN nodes. Subsequently, a message-passing operation is performed on the graph to generate node embeddings. These node embeddings are taken as the model weights for each RAN node for the upcoming FL round and are subsequently shared with the RAN nodes.

\section{Evaluation Metrics}
\label{sec:EvalMetrics}
\subsection{Anomaly Detection}
We employ commonly used metrics to measure the performance of anomaly detection models.
When training the model, we use the Mean Square Error (MSE) to compare the reconstructed time series to the observed time series.
For outlier detection, data points are classified into binary classes. Classification results are then compared to the annotated dataset using three metrics:
\begin{itemize}
    \item \textit{Precision}: $\frac{TP}{TP + FP}$ 
    \item \textit{Recall}: $\frac{TP}{TP + FN}$ 
    \item \textit{F1 Score}: $\frac{2.Precision . Recall}{Precision + Recall}$
\end{itemize}
where $TP$, $FP$ and $FN$ denote the number of respectively True Positives, False Positives and False Negatives.
\subsection{Federated Learning}
In FL, local data cannot be shared with a central server unlike in CL where all data is accessible centrally.  To assess the impact of data accessibility on training, we consider CL settings as a baseline for comparing various Federated Learning strategies due to its unrestricted data access. While the anomalies introduced during the annotation process in Section \ref{sec:dataCollection} attempt to replicate realistic scenarios, they are still a substitute for real-world settings. To address this issue, we propose utilizing the Centralized model to annotate previously unannotated data points in the dataset. 
To ensure a realistic evaluation in real-world scenarios, we use the classification results obtained on dataset 0 under Centralized settings as the reference for assessing classification performance. These data points, annotated during CL training, serve as our classification reference.
 Consequently, we assess the classification performance of FL experiments by measuring precision, recall, and F1-score metrics against the CL classification reference. Additionally, we assess the quality of our reconstruction results using the MSE. To quantify the communication efficiency gains achieved, we compare the data transmission requirements between FL and CL settings using $C_{footprint}$:
\[C_{footprint} = \frac{{(|\text{data communicated}|)}_{FL}}{{(|\text{data communicated}|)}_{CL}} = \frac{I. |\text{parameters}|}{|\text{data points}|}
\]
where $I$ represents the total number of iterations of FL rounds, $|\text{parameters}|$ the number of model parameters, and $|\text{data points}|$ the number of total training data points.
    
It is important to emphasize that evaluating personalized models on data other than their respective local data is irrelevant, as these models are designed to be exclusively deployed at the local level. Therefore, the evaluation of FL models will focus on assessing the central model using centralized data and assessing the personalized models using local data. To ensure convenience, the latter evaluation will be averaged across all clients, although it is possible to draw some distinctions between client types such as abnormal clients and normal clients, for instance.

\section{Experiments}
In this section, we present and discuss our findings. We begin by examining the outcomes of the centralized anomaly detection model. Next, we evaluate and compare the performance of the anomaly detection model trained using FL and CL under three experimental settings: normal behaviour across all cells,  and a geographic subset with numerous random anomalies.

\subsection{Centralized Anomaly detection}
\label{sec:centralized_learning}

We evaluate the following state-of-the-art models for multivariate time-series anomaly detection:
\begin{enumerate}
    \item LSTM \cite{wen2022transformers}: considered as the state-of-the-art model for time series forecasting.
    \item GConvLSTM and GConvGRU \cite{seo2018structured}: two Temporal Graph Neural Networks relying on the Graph Spectral Convolution for spatial features and the LSTM and GRU for time features.
    \item A3-GCN \cite{Zhu2020}: a Temporal Graph Neural Network relying on the attention mechanism.
\end{enumerate}

\subsubsection{Tuning of hyperparameters}
\label{sec:grid_search}

We conducted a grid search, as highlighted in Table \ref{tab:grid_search}, to identify the optimal combination of architectures and hyperparameters for the reconstruction task. This grid search was initially carried out on dataset 0 (described Table \ref{tab:dataset}) while considering graph-connected signals. Subsequently, a second experiment was conducted using independent signals. Due to computational limitations, the hyperparameter optimization was performed on the first 1000 time steps of the dataset.
\begin{table}[] 
\resizebox{\columnwidth}{!}{%
\begin{tabular}{|l|l|l|l|}
\hline
Parameters                                                                       & Optimization space & \begin{tabular}[c]{@{}l@{}} Connected \\TS\end{tabular}  & \begin{tabular}[c]{@{}l@{}} Independant \\ TS\end{tabular}              \\ \hline
\textbf{\begin{tabular}[c]{@{}l@{}}Recurrent\\ GNN\\ Architecture\end{tabular}}  & \begin{tabular}[c]{@{}l@{}}GConvLSTM \cite{seo2018structured}\\ GConvGRU \cite{seo2018structured}\\ A3-GCN \cite{Zhu2020}\end{tabular} & GConvLSTM  & GConvLSTM  \\ \hline
\textbf{Depth}                                                                   &\{1, 2, 3, 4, 5\}  & 2  & 1     \\ \hline
\textbf{History}                                                                 &\{16, 32, 48, 96, 192\} & 192 & 96 \\ \hline
\textbf{\begin{tabular}[c]{@{}l@{}}Nodes\\ embeddings\\ dimension\end{tabular}} &\{8, 16, 32, 64\} & 64 & 16           \\ \hline
\textbf{Learning rate}                                                           &{[}1e-4, 1e-1{]}   & 3e-3   & 3e-3                                                \\ \hline
\textbf{Batch size}                                                      &\{32, 64, 128\}  & 64  & 128                                                \\ \hline

\end{tabular}
}
\caption{Results of the optimization with regards to the Hyperparameters on the reconstruction tasks.}
\label{tab:grid_search}
\end{table}

\begin{figure}[tb]
\centering
  \includegraphics[width=\linewidth]{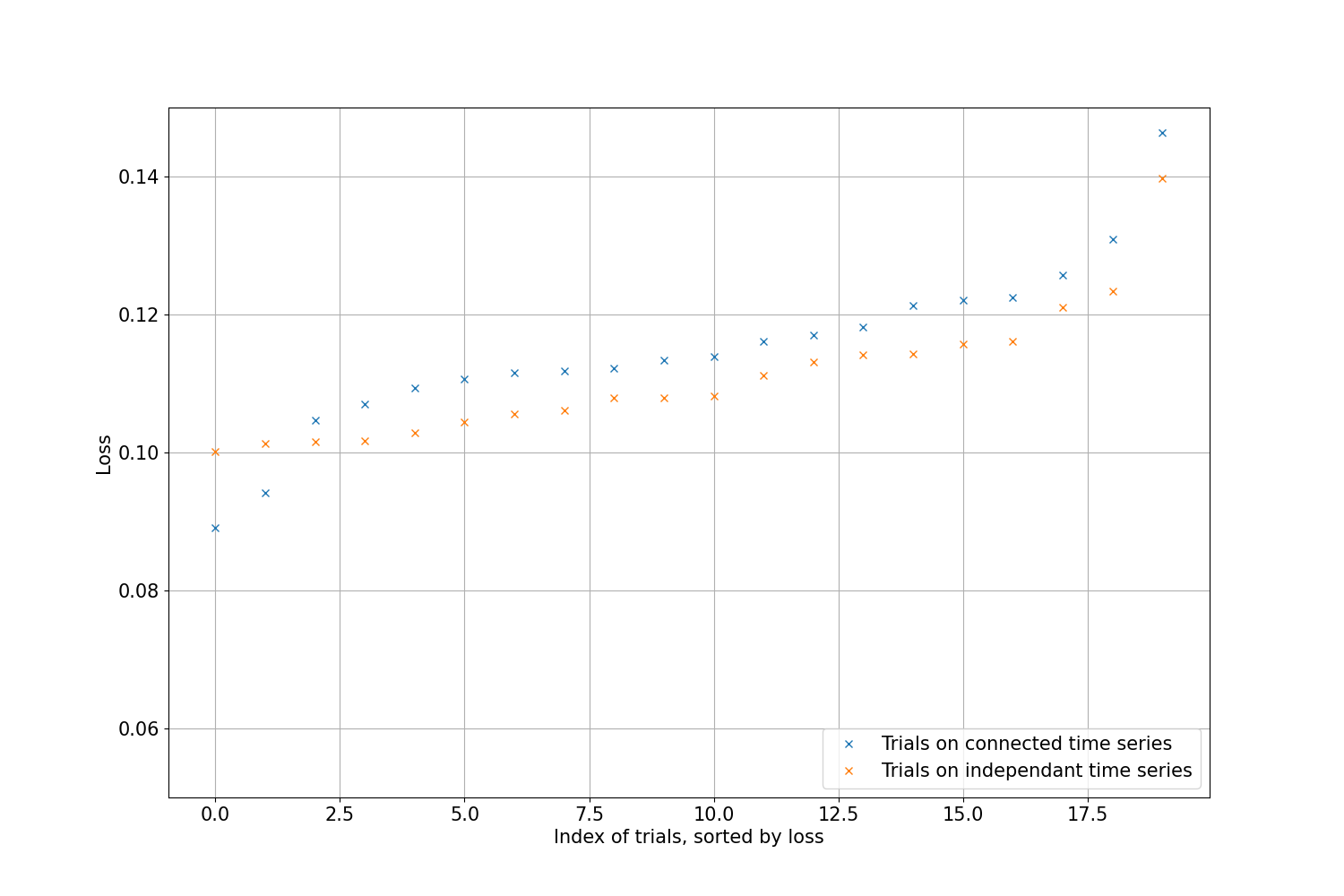}
  \caption{MSE loss of the different grid search trials, ranked from best to worst trial}
  \label{fig:grid_search_comparison}
\end{figure}

Figure \ref{fig:grid_search_comparison} depicts the outcomes of various grid search trials. It demonstrates that, following optimization, the GNN-based architecture yields superior reconstruction results, with a 10\% reduction in loss, when applied to graph-related data. These findings validate the advantages of integrating GNNs with the SW computational graph for modelling SW performance data.

\subsubsection{Performances with respect to different errors}

The anomaly detection method was assessed using two annotated datasets as described in \ref{sec:dataCollection}, and Table \ref{tab:dataset}. In each experiment, both the training and evaluation datasets were identical.  The main reason for this choice is that, while deploying the anomaly detection model in the network, faulty cells cannot be identified and removed beforehand. As such, under real settings, the model can possibly learn from RAN sites that experienced temporary problems, which is precisely what datasets 1 and 2 aim at emulating. 

We compared the results to the GConvLSTM architecture lacking signal connections, serving as our baseline to evaluate the impact of the SW graph. In this configuration, each node lacked neighbours, and graph convolution was independently applied to the node's data, followed by a single LSTM operation. This architecture can then be considered equivalent to a vanilla LSTM trained on the aggregated signal's data. 

For all experiments, we utilized the optimized parameters identified in Section~\ref{sec:grid_search}. The results can be found in Table \ref{tab:centralized_results}.

\begin{table}[]
\begin{tabular}{|l|l|l|l|l|l|}
\hline
Architecture & Dataset & MSE Loss   & Precision & Recall & F1     \\ \hline
LSTM               & 0      & 0.149 & -    & - & - \\ \hline
GNN-based               & 0      & 0.145 & -    & - & - \\ \hline
LSTM              & 1       & 0.182 & 0.4053    & 0.7850 & 0.5346 \\ \hline
GNN-based               & 1       & 0.179 & 0.4118    & 0.7850 & 0.5402 \\ \hline
LSTM               & 2       & 0.523  & 0.3913    & 0.4023 & 0.3967 \\ \hline
GNN-based                & 2       & 0.453  & 0.3970    & 0.4023 & 0.3996 \\ \hline

\end{tabular}
\caption{Reconstruction and classification performances on the two annotated datasets}
\label{tab:centralized_results}
\end{table}

The developed method exhibited varying success in detecting anomalies. Analyzing different metrics, we observe the following:
\begin{itemize}
    \item \textbf{Improvement with GNNs:} The use of Graph Neural Networks (GNNs) led to performance improvements, albeit with different degrees of success. On datasets 0 and 1, GNNs showed slight enhancements compared to the baseline LSTM method. However, the gains in reconstruction performance were notably more significant in dataset 2. Notably, dataset 2 involved the propagation of anomalies to other data points, highlighting that GNNs outperformed non-graph methods in this scenario.
    \item \textbf{Similar Classification Performance:} The classification performance of the GNN-based architecture closely resembled that of the LSTM baseline. Although there were slight improvements in the precision metric, GNNs did not yield substantial classification benefits.
    \item \textbf{Average Precision:} The overall precision score across all experiments was approximately 40\%, indicating that 60\% of identified anomalies were not artificially introduced but stemmed from the real data. This limitation arises from the challenge of constructing an annotated dataset for method evaluation without prior knowledge of anomalies. It is expected that more data points (False Positives) will be flagged as anomalies. Importantly, the number of False Positives was comparable to the number of anomalies introduced in the dataset following Alg. \ref{alg:data_annotation}, aligning with domain experts' expectations regarding abnormal data points. This suggests that the precision score validates the anomaly detection method.
    \item \textbf{Recall Performance:} The relatively high recall score for dataset 1 indicates that the method effectively detected introduced anomalies, validating its effectiveness. However, for dataset 2, there was a substantial drop in the recall score. This, coupled with a significant increase in loss (2.5x), indicates that the method struggled to perform well on data with propagated anomalies, making it challenging to validate the method's performance in this context
\end{itemize}

\subsection{Federated Anomaly detection }

Various FL strategies have been run and compared under different settings. 

\subsubsection{Performances on normal data}
\label{sec:FL_normal_data}
We first simulate FL training using normal data. We evaluate FL detections by comparing them to the detection results on dataset 0, which serves as the ground truth. 
We maintain a fixed total number of 100 epochs for FL (in contrast to the CL baseline with 50 epochs) while exploring various aggregation frequencies, referred to as local epochs. We adjust the number of iteration rounds, denoted as $I$, to ensure a consistent total number of epochs. For clarity and brevity, we name each experiment as "Aggregation $I$x$local\_epoch$" (e.g., "FedAvg 5x20" signifies the FedAvg aggregation rule with 5 rounds and 20 local epochs). Results are summarized in Table \ref{tab:FL_results_normal}.

\begin{table}[]
\resizebox{\columnwidth}{!}{%
\begin{tabular}{|l|l|l|l|l|}
\hline
 FL Strategy & \begin{tabular}[c]{@{}l@{}}Personalized models\\ performance\\ loss/precision/recall/f1\end{tabular} & \begin{tabular}[c]{@{}l@{}}Central model\\ performance\\ loss/precision/recall/f1\end{tabular} & $C$ \\ \hline
FedAvg-10x10                                                              & -                                                                                                     & 0.226/0.719/0.707/0.695     &      7.84 \%                                                            \\ \hline
FedAvg-20x5                                                             & -                                                                                                     & 0.198/0.724/0.707/0.699          & 15.68 \%                                                              \\ \hline
FedAvg-5x20                                                               & -                                                                                                     & 0.281/\textbf{0.777}/0.677/0.701     & 3.92 \%                                                                  \\ \hline
 FedAvgReg-5x20                                                               & -                                                                                                     & 0.292/0.733/0.702/0.691  &  3.92 \%                                                                \\ \hline
 FedGraph-10x10                                                              & 0.219/0.713/0.702/0.686                                                                               & 0.219/0.712/0.702/0.686        &   7.84 \%                                                            \\ \hline
 FedGraph-20x5                                                                & \textbf{0.197}/0.723/\textbf{0.726}/\textbf{0.708}                                                                               & \textbf{0.197}/0.723/\textbf{0.726}/\textbf{0.708} & 15.68 \%                                                                    \\ \hline
 FedGraph-5x20                                                               & 0.269/\textbf{0.760}/0.665/0.683                                                                               & 0.276/0.757/0.667/0.683   & 3.92 \%                                                                     \\ \hline
 FedGraphReg-5x20                                                               & 0.275/\textbf{0.760}/0.646/0.672                                                                               & 0.282/0.758/0.645/0.669   & 3.92 \%                                                                     \\ \hline
\end{tabular}
}
\caption{FL performances compared to the centralized training baseline. \textbf{Bold} figures represent the best values for each metric. As a comparison, the loss achieved with centralized training was 0.145.}

{\addvspace{-1\baselineskip}}
\label{tab:FL_results_normal}
\end{table}

In all experiments, FL training methods exhibited high classification performance, indicating the comparability of FL and CL training. Interestingly, despite differences in reconstruction metrics, the experiments demonstrated similar classification metrics, suggesting that complex fitting models are not necessary for identifying most anomalies

While looking at the reconstruction performances, it can be noted that:
\begin{itemize}
    \item \textit{FedGraph} aggregation outperformed \textit{FedAvg} slightly.
    \item The introduction of a regularization factor in the loss resulted in a slight performance degradation.
    \item These modifications had minimal impact compared to aggregation frequency. Frequent aggregations, such as low local epochs, consistently provided the best reconstruction improvements across all strategies.
\end{itemize}

Furthermore, in all experiments, FL settings resulted in substantial communication savings compared to CL. However, these communication gains were directly tied to less frequent aggregations, which led to poorer fitting performance.

\begin{figure*}[]
    \centering
    \begin{subfigure}[b]{0.49\textwidth}
      \centering
      \includegraphics[width=\textwidth]{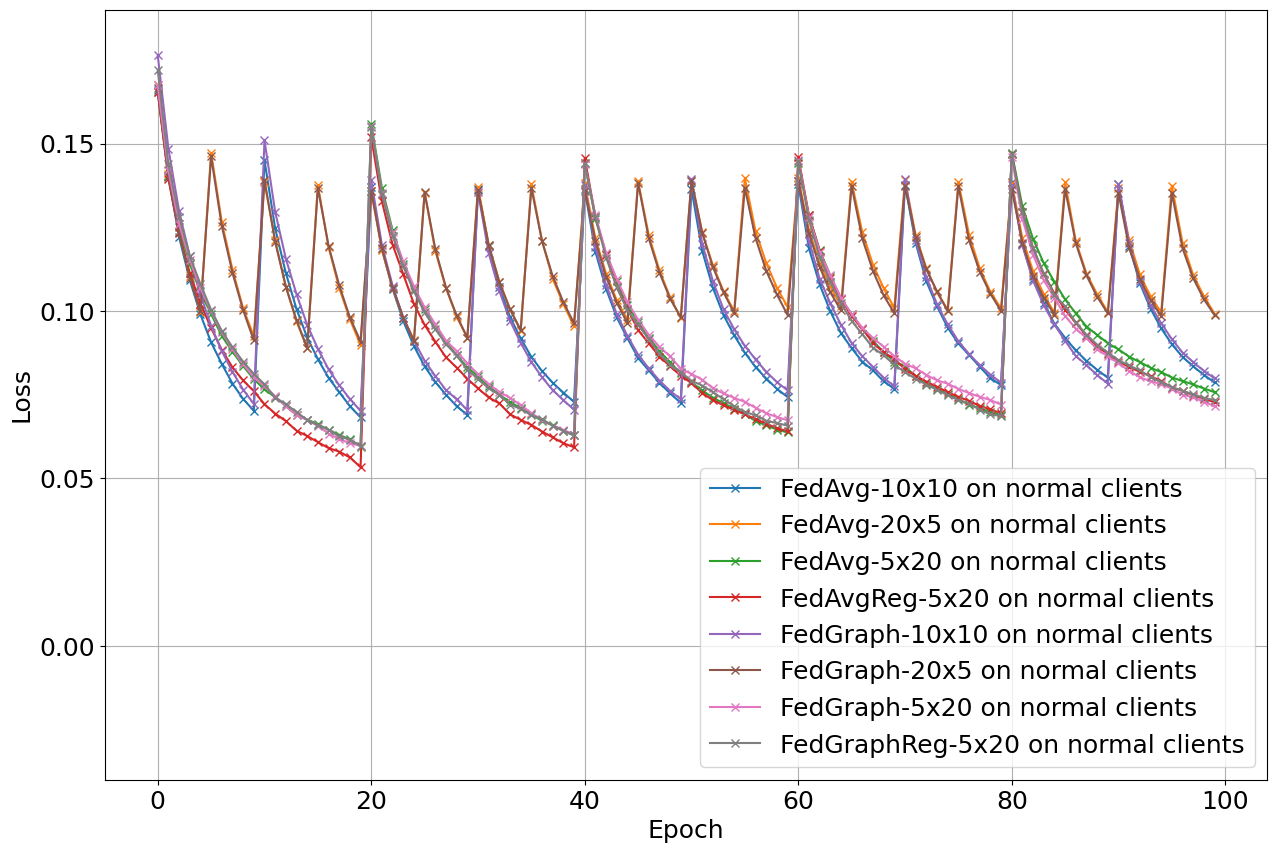}
      \caption{Loss}
      \label{fig:normal_loss}
    \end{subfigure}%
    \hfill
    \begin{subfigure}[b]{0.49\textwidth}
      \centering
      \includegraphics[width=\textwidth]{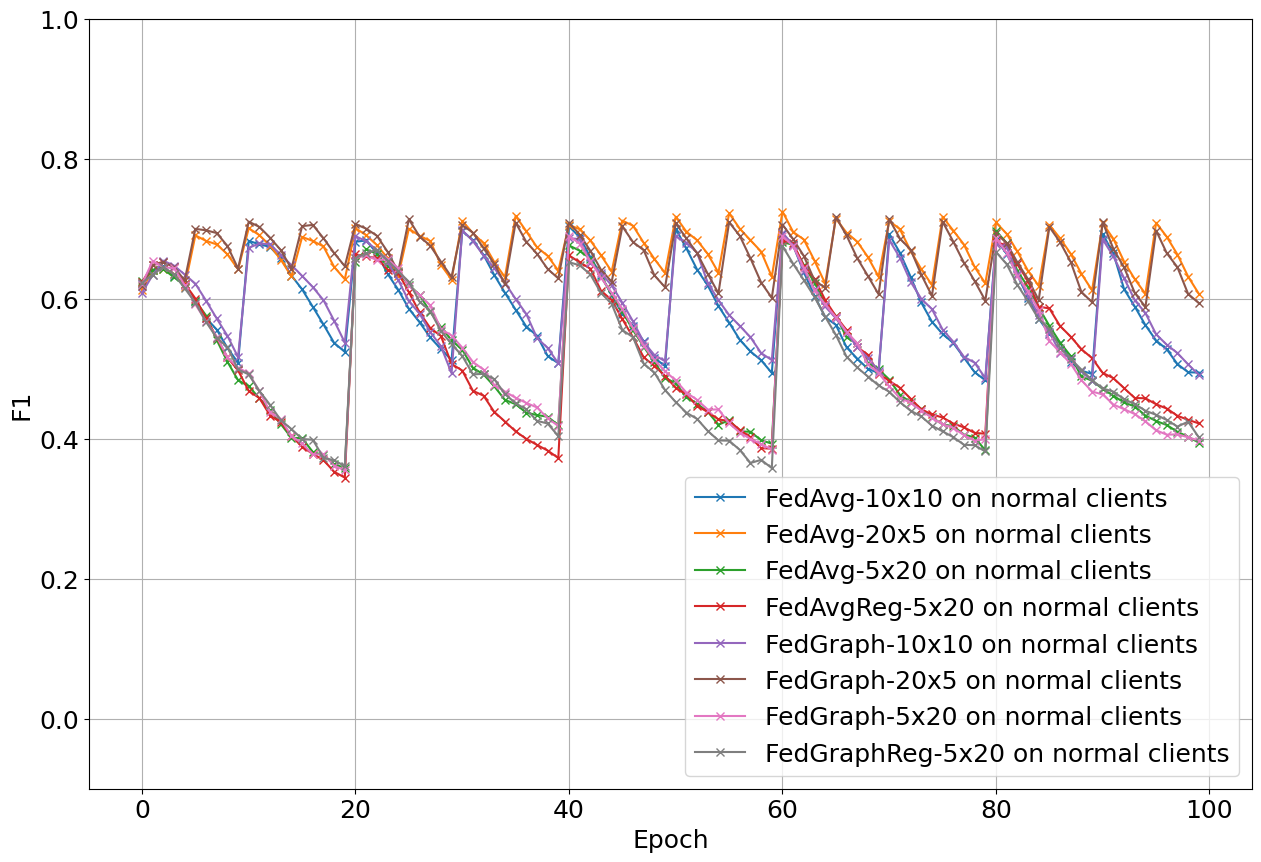}
      \caption{F1}
      \label{fig:normal_f1}
    \end{subfigure}
\caption{Performance metrics of the different experiments during FL training on normal data}
\label{fig:normal_metrics}
\end{figure*}

\begin{figure}
    \centering
    \includegraphics[width =\linewidth]{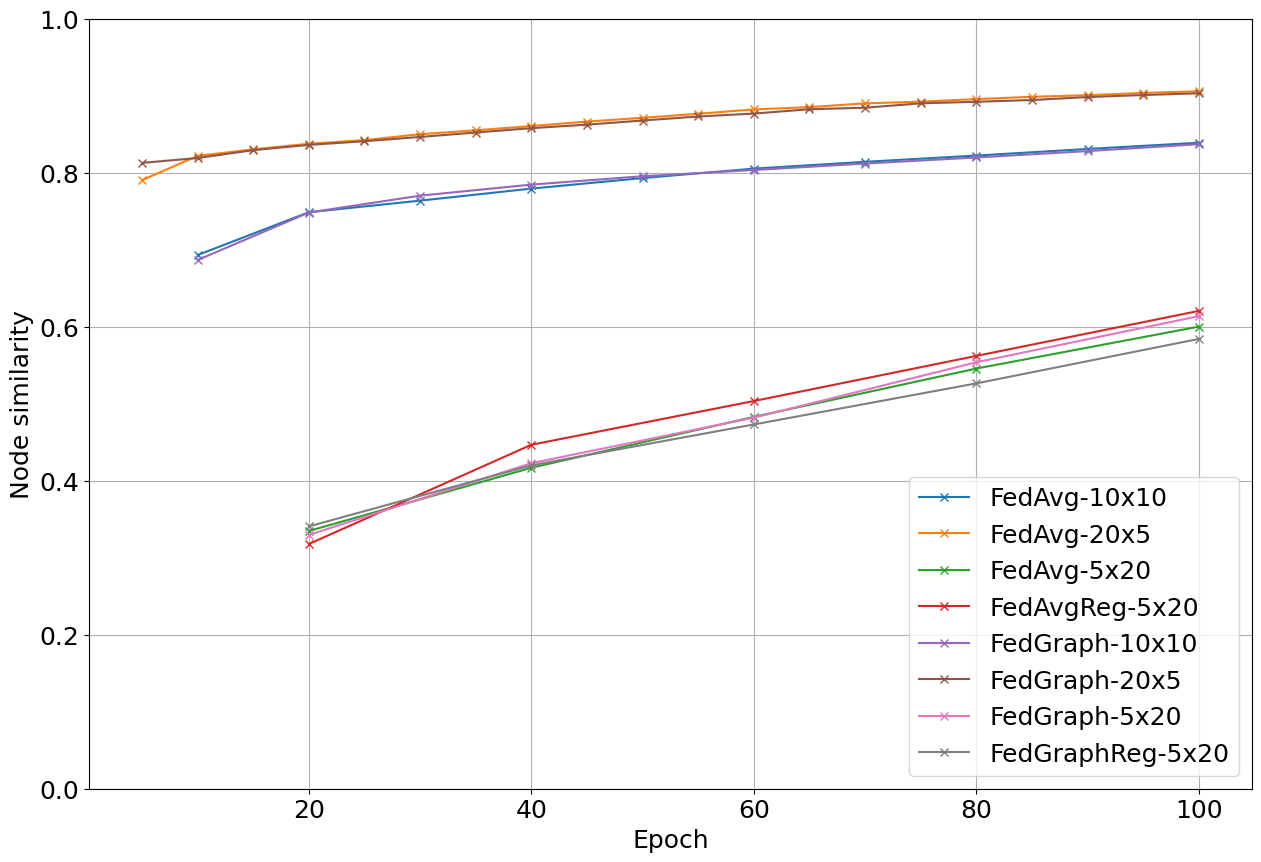}
    \caption{Evolution of the similarities between local models throughout the FL training on normal data}
    \label{fig:normal similarity}
\end{figure}

Figure \ref{fig:normal_metrics} illustrates the evolution of performance metrics during FL training. These plots reveal that FL training is more susceptible to overfitting compared to CL training. While local models improved their fit to local data during local training, they simultaneously experienced rapid degradation in classification performance. A similar trend can be observed in the similarity between locally trained models throughout the training, as depicted in Figure \ref{fig:normal similarity}. Longer local epochs led to less similarity among local models, indicating overfitting to local data. Frequent central aggregations emerged as a valuable regularization tool to ensure the success of FL training. While the \textit{FedGraph} aggregation rule initially provided slight improvements, it later exacerbated overfitting in later FL rounds, resulting in less similar models across cells. The significance of this personalization mechanism remains unclear.

In summary, our experiments demonstrated that our settings did not require extended FL training to achieve good performance, as rapid convergence of metrics was observed after the first round.

\subsubsection{Performances on abnormal dataset}
\label{sec:FL_abnormal_data}
To study FL performance in the presence of anomalous cells within the network, we conduct FL training using dataset 1. We then assess FL detections by comparing them to the ground truth classifications obtained from CL detection on dataset 1. We summarize the results in Table \ref{tab:FL_results_abnormal}.

\begin{table}[]
\resizebox{\columnwidth}{!}{%
\begin{tabular}{|l|l|l|l|}
\hline
FL Strategy & \begin{tabular}[c]{@{}l@{}}Personalized models\\ performances\\ loss/precision/recall/f1\end{tabular} & \begin{tabular}[c]{@{}l@{}}Central model\\ performances\\ loss/precision/recall/f1\end{tabular} & $C$\\ \hline
FedAvg-5x20                                                               & -                                                                                                     & 0.292/0.085/0.970/0.114 & 3.92 \%                                                                        \\\hline

FedAvgReg-5x20                                                               & -                                                                                                     & \textbf{0.288}/\textbf{0.088}/0.971/\textbf{0.117 } & 3.92 \%                                                                        \\ \hline
FedGraph-5x20                                                               & 0.312/\textbf{0.077}/\textbf{0.985}/\textbf{0.106}                                                                           & 0.319/0.077/\textbf{0.985}/0.106  & 3.92 \%                                                                      \\ \hline
FedGraphReg-5x20                                                               & \textbf{0.291}/0.069/0.978/0.094                                                                               & 0.297/0.069/0.978/0.093  & 3.92 \%                                                                    \\ \hline
\end{tabular}
}
\caption{FL performances compared to the centralized training baseline. \textbf{Bold} figures represent the best values for each metric. As a comparison the loss achieved with centralized training was 0.179.}
{\addvspace{-1.5\baselineskip}}
\label{tab:FL_results_abnormal}
\end{table}

\begin{figure*}[]
\centering
\begin{subfigure}[b]{0.49\textwidth}
  \centering
  \includegraphics[width=\textwidth]{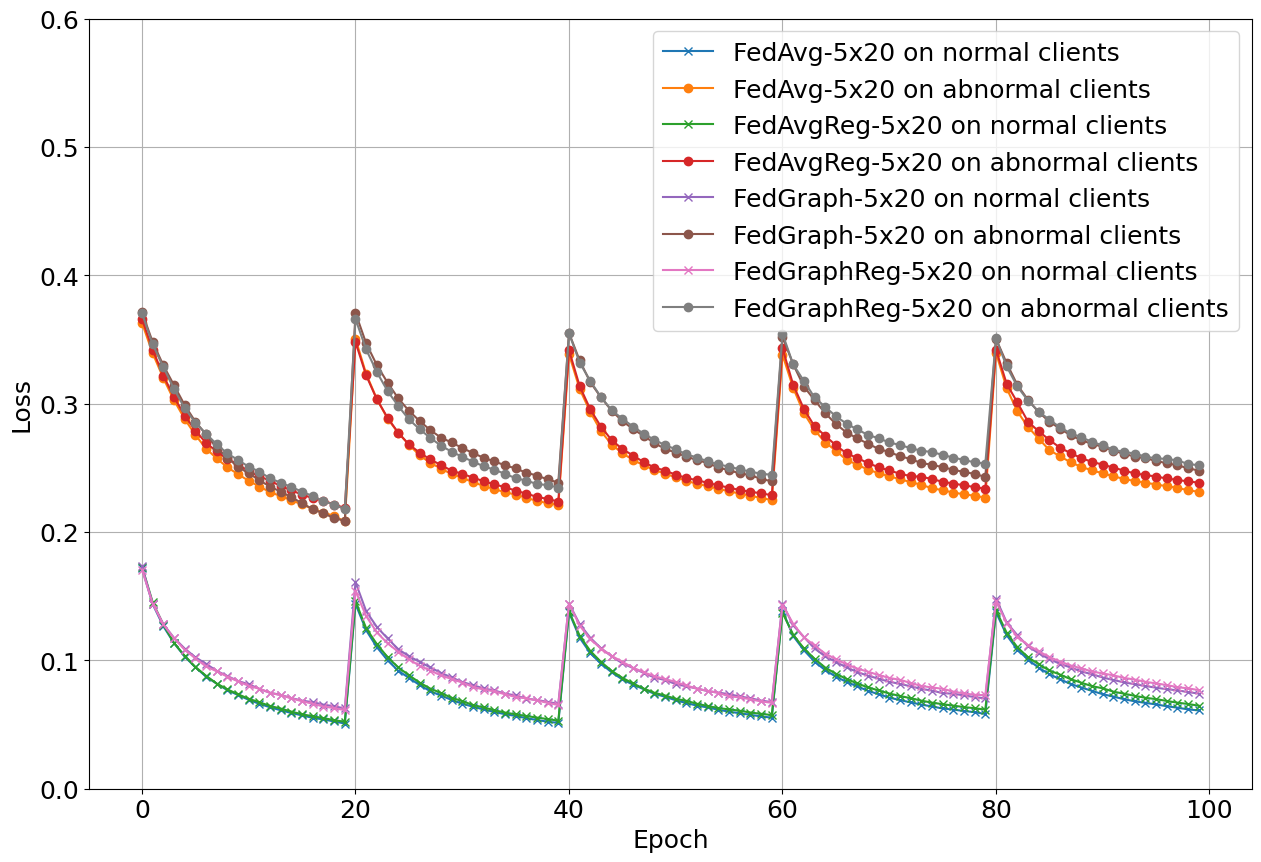}
  \caption{Loss}
  \label{fig:abnormal_loss}
\end{subfigure}%
\hfill
\begin{subfigure}[b]{0.49\textwidth}
  \centering
  \includegraphics[width=\textwidth]{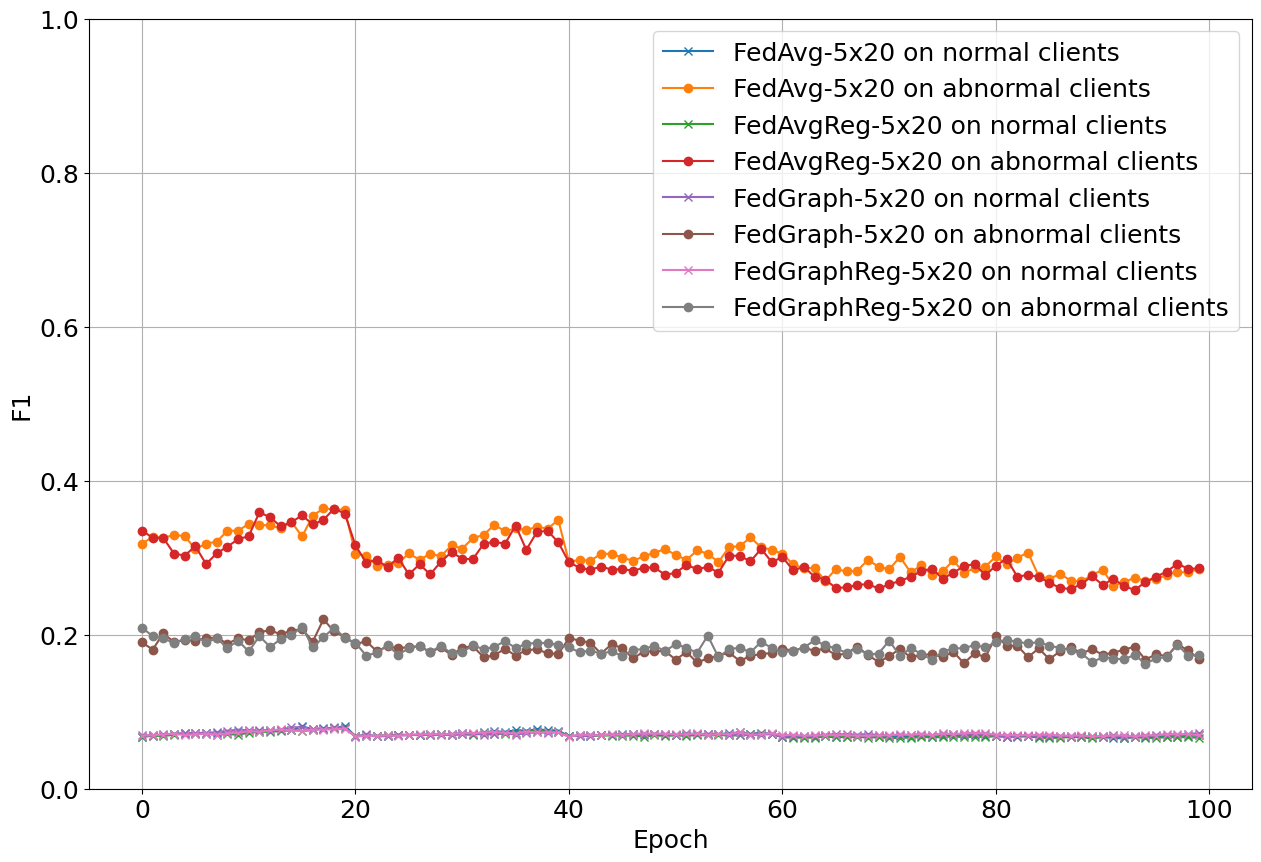}
  \caption{F1}
  \label{fig:abnormal_f1}
\end{subfigure}
\caption{Performance metrics of the different experiments during FL training on normal cells and abnormal cells.}
\label{fig:abnormal_metrics}
\end{figure*}

All experiments consistently yielded subpar classification results with high recall but low precision, indicating an excessive detection of anomalies. Overall, FL performed inadequately on this dataset, highlighting its significant inferiority compared to CL in the context of abnormal cells.
Turning our attention to reconstruction performance, the following observations can be made:
\begin{itemize}
    \item The personalized aggregation degraded both the reconstruction and classification performance.
    \item The incorporation of regularized loss contributed to improved fitting results.
\end{itemize}

\begin{itemize}
    \item Abnormal clients exhibited a predominant share of the reconstruction error but achieved superior detection performance.
    \item The \textit{FedGraph} aggregation rule significantly degraded detection performance on abnormal cells while maintaining performance on normal cells comparable to \textit{FedAvg}.
    \item Despite rapid improvements in fitting performance, classification performance remained relatively stable and low. In these experimental conditions, FL training did not match the performance of CL learning.
\end{itemize}

\begin{figure*}[]
\centering
\begin{subfigure}[b]{0.33\textwidth}
  \centering
  \includegraphics[width=\textwidth]{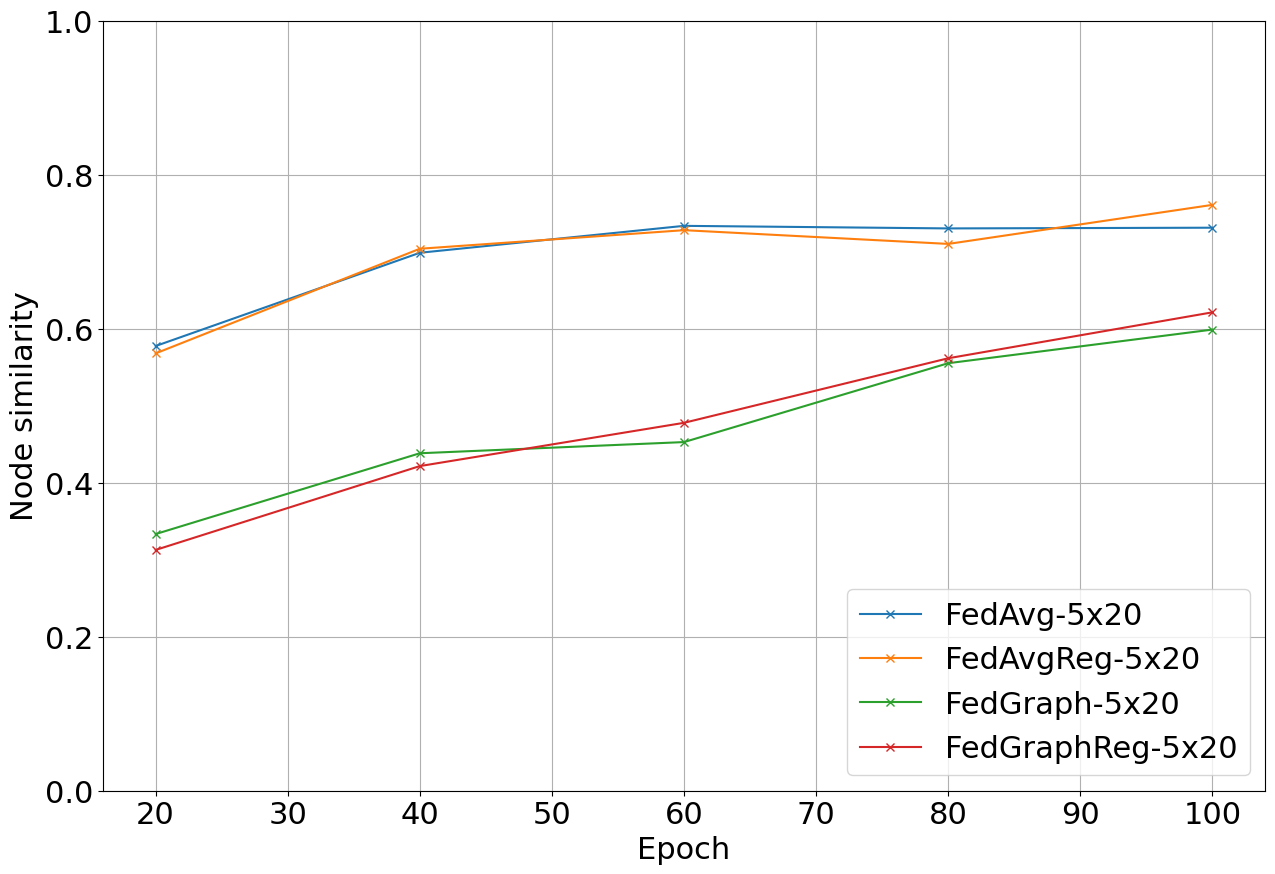}
  \caption{Between normal cells}
  \label{fig:normal_sim}
\end{subfigure}%
\begin{subfigure}[b]{0.33\textwidth}
  \centering
  \includegraphics[width=\textwidth]{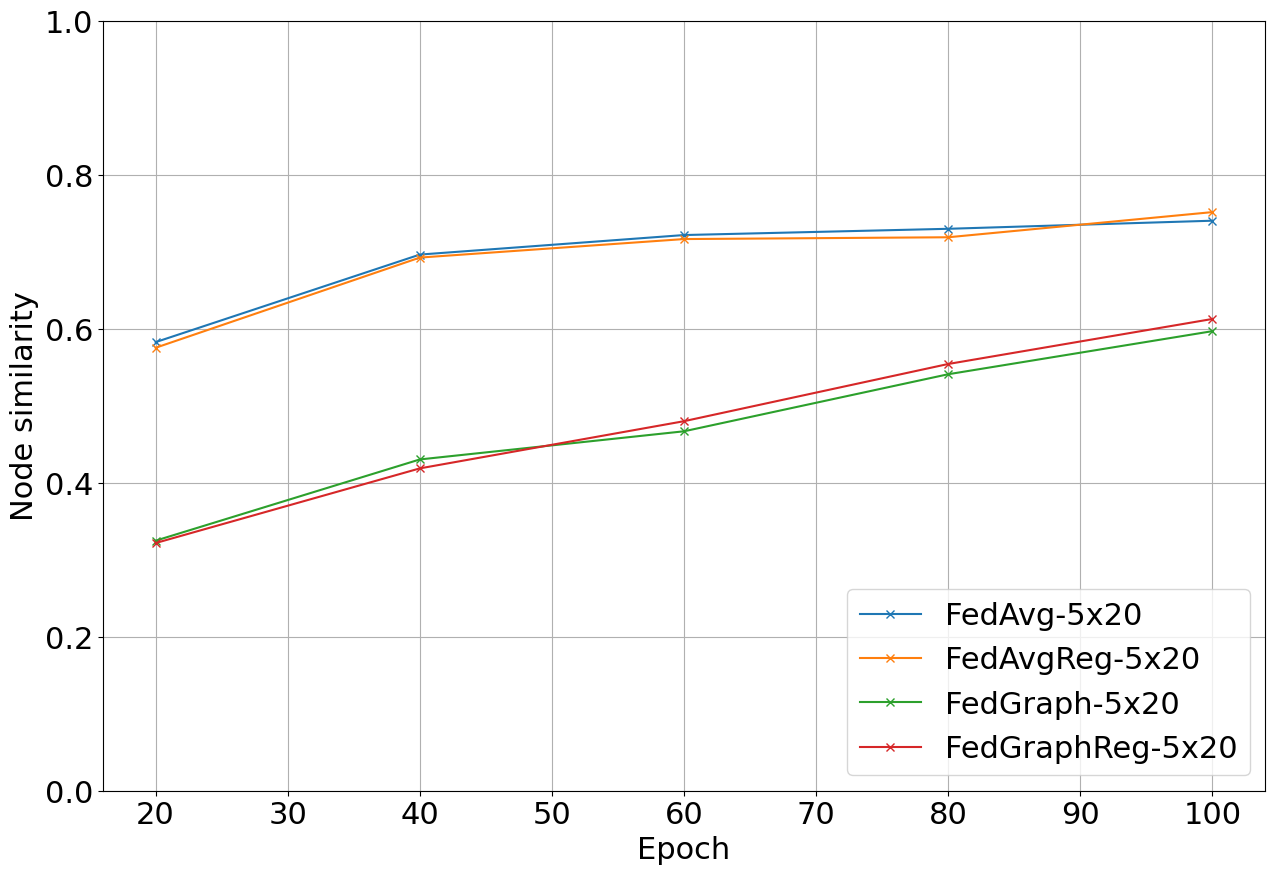}
  \caption{Between normal and abnormal cells}
  \label{fig:mixed_sim}
\end{subfigure}%
\begin{subfigure}[b]{0.33\textwidth}
  \centering
  \includegraphics[width=\textwidth]{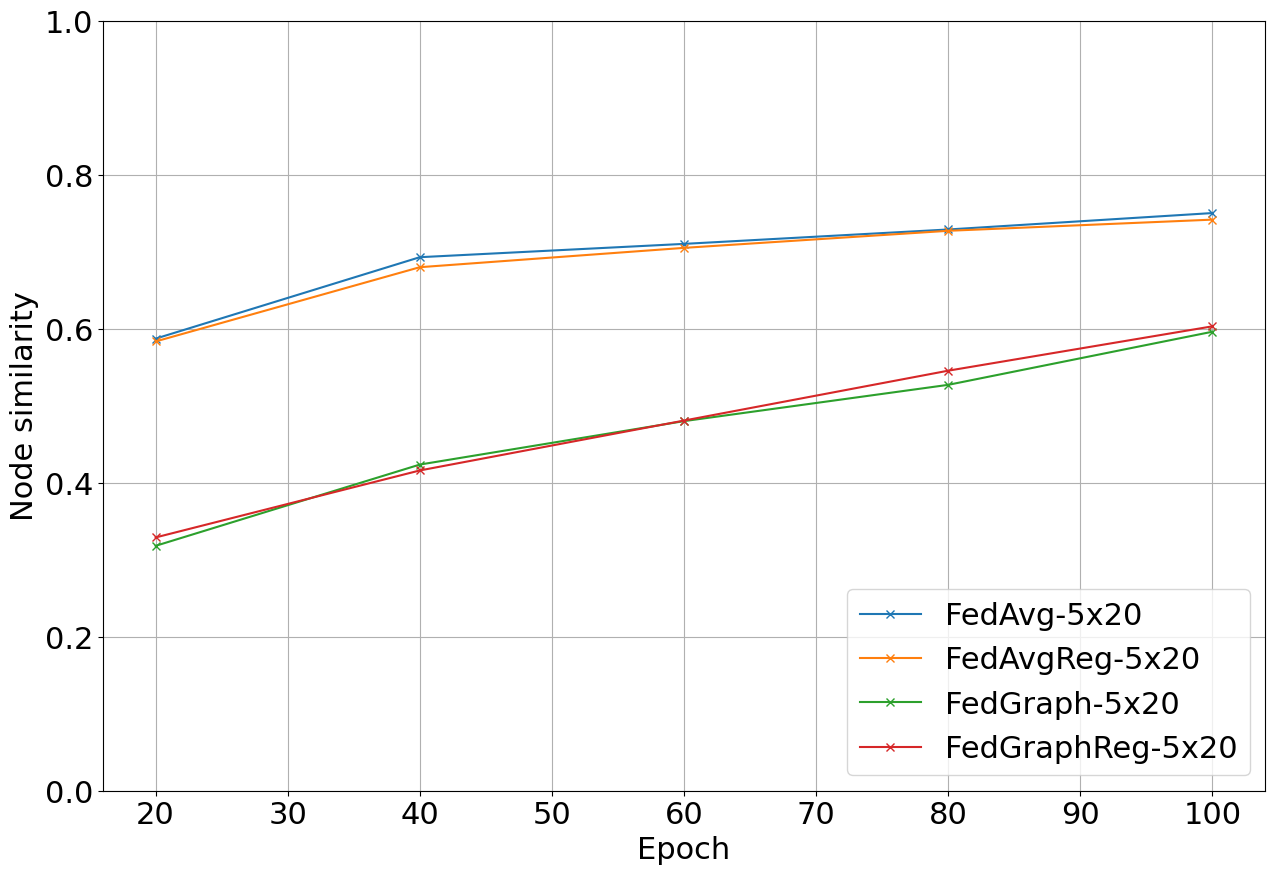}
  \caption{Between abnormal cells}
  \label{fig:abnormal_sim}
\end{subfigure}
\caption{Evolution of the average link weight between local cells throughout the FL training on dataset 1.}
{\addvspace{-0.5\baselineskip}}
\label{fig:abnormal_similarity}
\end{figure*}

Furthermore, when examining the similarities between the local models returned by the clients (refer to Figure \ref{fig:abnormal_similarity}), we observe an intriguing trend. In comparison to training on normal data (refer to Figure \ref{fig:normal similarity} in Section \ref{sec:FL_normal_data}), our personalized aggregation approach resulted in significantly less similarity among models, indicating a higher level of personalization. Moreover, this personalization effect was consistent across various cell types, as demonstrated in Figures \ref{fig:normal_sim}, \ref{fig:mixed_sim}, and \ref{fig:abnormal_sim}. Connecting these findings to the metrics presented in Table \ref{tab:FL_results_abnormal}, it becomes evident that the personalization process negatively impacted reconstruction performance and, to a lesser extent, detection performance.

\section{Discussion}


The centralized anomaly detection task yields intriguing results. While GNNs improves fitting performance significantly, detection performance remains akin to the baseline. This can be attributed to the obvious nature of anomalies in the evaluation datasets, rendering top-notch fitting results unnecessary for effective detection. The CL framework further exacerbates this effect. In CL, the model detects anomalies using shared knowledge from all datasets, leading it to prioritize general fitting models and discouraging excessive model personalization. This duality is advantageous as it demonstrates the method's robustness to obvious anomalies and its capacity for generalization. However, it is also disappointing as it does not allow us to evaluate the method's ability to identify "silent issues" that can only be discerned by exploiting the connections between time-series measurements rather than monitoring them in isolation.

This outcome was expected for dataset 1 (refer to Table \ref{tab:dataset} for dataset description) since the anomalies in this dataset were introduced without any signal interdependencies. However, for dataset 2, the results may seem counterintuitive, as the introduced anomalies were meant to propagate to other signals. In this regard, despite the significant improvements in fitting results, which initially appeared promising, we anticipated that the GNN-based method would perform better on the classification task by leveraging the graph connections between signals. Yet, upon closer examination of dataset 2, we observe two key factors contributing to this unexpected outcome.

First, the propagation of anomalies in dataset 2 resulted in a substantial increase in the total number of anomalies compared to dataset 1 (a fivefold increase). Consequently, this dataset presents a greater challenge, as a reconstruction-based anomaly detection method like ours, designed to "learn" the normal data behaviour, is expected to perform best when the training data contains fewer anomalies.

Second, an alternative explanation for this decline in performance is that when anomalies are propagated to different data points, they may have originated from sources with varying scales. As a result, these propagated anomalies can become indistinguishable from the inherent noise of time-series measurements with larger scales. 

If we focus on the FL experiments, the failure of the FL training in the context of abnormal cells is rather unexpected. Initially, FL was anticipated to handle abnormal client data effectively through shared knowledge, as seen in normal cell scenarios with frequent aggregations. However, FL exhibited opposing behaviours in these two contexts. Notably, the robustness observed in normal cell models, where variable fitting results did not hinder performance, is not evident in abnormal settings; all strategies fail equally.

This limitation of FL in abnormal settings becomes even more evident when we consider that the CL anomaly detection model was validated using the same training data. One explanation for this behaviour is that the fitting performances were just too low and that overall the lack of accurate fitting led to many data points to deviate from the reconstruction and as such that an outstanding number of False Positives are observed.
Another explanation for this behaviour is overfitting during local training. Under CL settings, the mix of normal and abnormal data acted as a regularizer, helping the model learn the "normal behaviour" of the data. Conversely, in FL settings, local models for abnormal cells inadvertently fit abnormal data, which contradicts the goal of a reconstruction-based anomaly detection method (relying on the model learning the "normal behaviour" of data). Interestingly, in this context, the aggregation of abnormal and normal models in FL training does not resemble the aggregation of normal and abnormal data in CL training. Consequently, the limitations of our FL anomaly detection model for this dataset become evident.

It is worth mentioning that FL preserves local data privacy; however, recent discussions have raised concerns about privacy attacks,  particularly Deep Leakage \cite{zhu2019deep} in FL settings. Future studies should investigate the robustness of FL in Telecom settings to better understand potential privacy vulnerabilities and safeguards against them.

Our motivation for utilizing Personalized FL, particularly Graph-based FL, was to provide models that align better with local conditions for cells deployed under similar settings and scenarios (e.g., similar traffic, weather conditions, etc.). Once again, we observe divergent effects of personalized aggregation in the two simulated scenarios. In normal cells, FedAvg yielded modest improvements in reconstruction and classification, while in abnormal cell scenarios, FedAvg underperformed.

However, in the case of abnormal cells, the aggregation process yielded significantly more personalized models, irrespective of cell type (normal or abnormal). While increased personalization in abnormal cells was expected due to data deviating significantly from "normal" data, it remains unclear why normal cells were also impacted by this personalization process. It was anticipated that normal cells would exhibit higher similarities to each other and less similarity with abnormal cells, but our observations show comparable similarities regardless of cell type.

One possible explanation for this phenomenon is that the anomalies introduced in abnormal cells essentially constitute noise (see the method for annotation in algorithm \ref{alg:data_annotation}). The local models trained within abnormal cells represent a noisier version of the "average model." Consequently, the local models trained within abnormal cells represent a noisier version of the local models. In this context, the concentration of noise in 20\% of the cells seemed substantial enough to pull the global model away from the global optimum.

Interestingly, the two aggregation strategies handled these constraints differently. FedAvg acts as a regularization factor, averaging all models and gradually approaching the average model. Conversely, FedGraph performs a weighted average that discourages aggregations with other cell models, hindering the denoising of local models through shared aggregations. It would be intriguing to investigate whether these different strategies ultimately converge toward a common model and if personalized aggregation on noisy cells is simply a slower-converging variation of the FedAvg aggregation, that could be paced up for example with more frequent aggregations.

\section{Conclusions}

We present a Federated GNN-based anomaly detection model 
that can detect and diagnose anomalies for multivariate telecom data.
Specifically, the model achieves an improvement when modelling the combination of SW execution graph and telecom network topology using bi-level temporal GNNs. Moreover, federated learning allows it to be able to leverage data while addressing privacy concerns. 
The method makes it a good choice for telecom systems where accurate anomaly detection is required, and also privacy constraints need to be addressed.

Throughout this work, we faced several challenges. There is a lack of benchmark datasets for fault detection tasks in telecom networks. As technology powering networks evolves constantly, it is a challenge to obtain an annotated dataset. Therefore, we relied in this work on unsupervised learning. Because of the difficulties to create a method for annotating an evaluation dataset, the industrial reliability of the reconstruction-based anomaly detection method developed can be improved. In practical settings, telecom operators often run field trials to assess the reliability of such methods. Moreover, FL showed comparable results to CL training while minimizing communication costs. In some cases, FL did not attain desirable results. The choice of a reconstruction-based anomaly detection model appeared to be really challenging to train. In the future, we propose to explore techniques to amplify errors and improve training stability. In most cases, a strategy relying on frequent aggregations is to be preferred to prevent high biases towards fitting local data, at the expense of a higher volume of data communicated. Additionally, despite some slight improvements, the FedGraph aggregation strategy appeared to be comparable to FedAvg for our assessment data. The graph-based personalization appeared to be promising but further large-scale trials should be run to assess the real impact of this technique. 
It is worth noting that FL provided significant communication gains for our settings. The FL training also appeared to converge rapidly which may allow further communication gains.

To the best of our knowledge, this is the first work that investigates the intertwined SW execution flow and telecom network deployment under federated learning settings. In the future, we propose to explore the limitations we highlighted with the stability of FL, and expand the analysis to incorporate diverse temporal trends and deployment scenarios. 

\section{Acknowledgment}

We sincerely thank Prof. Aristides Gionis for his feedback and his support during the project.

\bibliographystyle{./IEEEtran}
\bibliography{bibliography}

\end{document}